\documentclass[runningheads]{llncs}
\usepackage{graphicx}

\usepackage[dvipsnames]{xcolor} %

\usepackage{tikz}
\usepackage{comment}
\usepackage{amsmath,amssymb} %
\usepackage{color}

\newcommand\B[1]{\textcolor{blue}{#1}}

\newcommand\BL[1]{\textcolor{blue}{#1}}
\newcommand\RE[1]{\textcolor{red}{#1}}

\newcommand\OR[1]{\textcolor{orange}{#1}}
\definecolor{newgreen}{rgb}{0, 0.6, 0.2}
\newcommand\GR[1]{\textcolor{newgreen}{#1}}

\usepackage{amsmath}
\usepackage{bm}

\usepackage{dutchcal} %

\usepackage{algorithm}
\usepackage{algpseudocode}

\usepackage[hang,flushmargin]{footmisc} %

\usepackage{graphicx}
\usepackage{wrapfig} %
\usepackage{amssymb}

\usepackage{booktabs} %
\usepackage{array}
\usepackage{makecell} %
\usepackage{arydshln} %
\usepackage{multirow}
\usepackage{threeparttable} %
\usepackage{enumitem}
\usepackage[pagebackref=true,breaklinks=true,colorlinks,bookmarks=false,citecolor=NavyBlue,linkcolor=red]{hyperref}

\newcommand\bb[1]{\textbf{#1}}

\newcommand\x{$\times$}

\usepackage{pifont} %

\newcommand\ie{\textit{i.e.}}
\newcommand\eg{\textit{e.g.}}
\newcommand\etal{\textit{et al.}}
\newcommand\vs{\textit{vs.}}

\begin{document}
\pagestyle{headings}
\mainmatter

\title{R2L: Distilling Neural \textit{Radiance} Field to Neural \textit{Light} Field for Efficient Novel View Synthesis}

\titlerunning{R2L: Distilling NeRF to NeLF for Efficient NVS}
\author{Huan Wang\inst{1,2,\ast}\and
Jian Ren\inst{1,\dag} \and
Zeng Huang\inst{1,\ddag} \and
Kyle Olszewski\inst{1}\and
Menglei Chai\inst{1} \and
Yun Fu\inst{2} \and
Sergey Tulyakov\inst{1}}
\authorrunning{H. Wang et al.}
\institute{Snap Inc.\and Northeastern University, USA\\
\href{https://snap-research.github.io/R2L}{Project: https://snap-research.github.io/R2L}}

\renewcommand{\thefootnote}{\fnsymbol{footnote}}
\footnotetext[1]{Work done when Huan was an intern at Snap}
\footnotetext[2]{Corresponding author: jren@snapchat.com}
\footnotetext[3]{Now at Google}

\maketitle

\begin{figure}
    \centering
    \begin{tabular}{c@{\hspace{0.01\linewidth}}c@{\hspace{0.01\linewidth}}c}
     \includegraphics[width=0.48\linewidth]{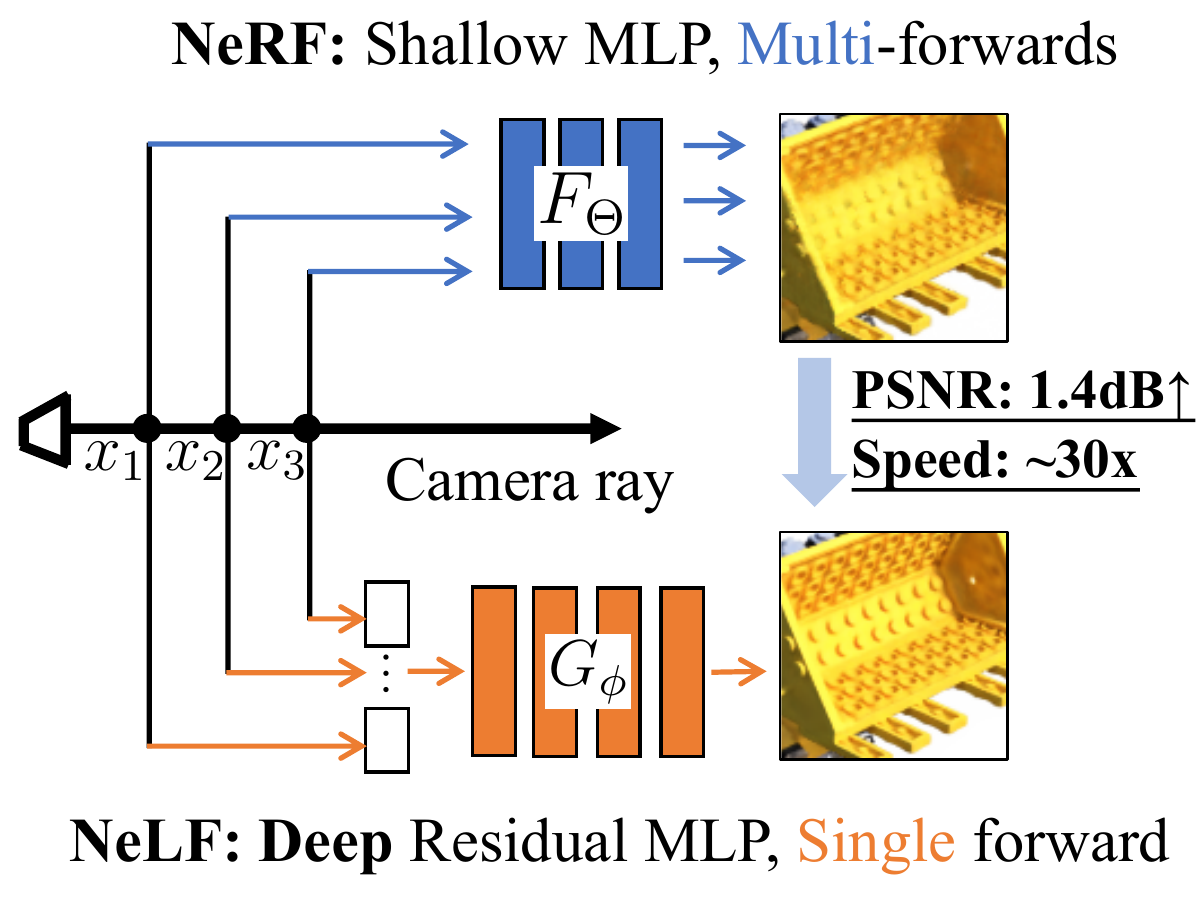} &
    \includegraphics[width=0.48\linewidth]{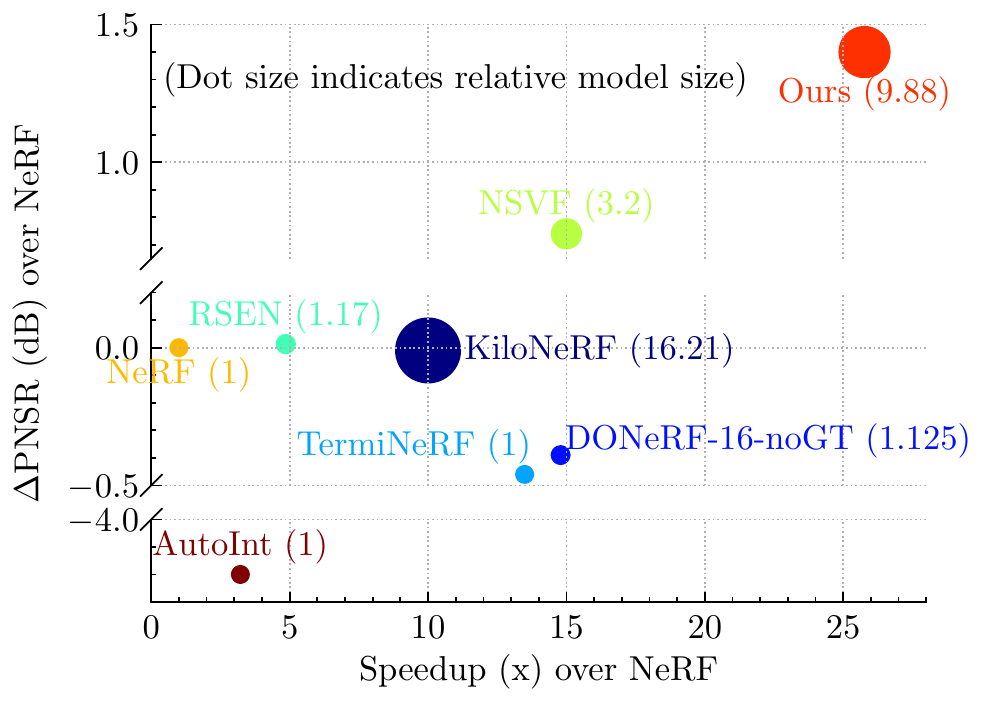} \\
    (a) NeRF~\vs~our NeLF method & (b) Speedup-PSNR-Model Size comparison
    \end{tabular} 
    \caption{(a) Our neural light field (NeLF, bottom) method improves the rendering quality by $1.40$ PSNR over neural radiance field (NeRF, top)~\cite{mildenhall2020nerf} on the NeRF synthetic dataset,  while being around $30\times$ faster. (b) Our method achieves a more favorable speedup-PSNR-model size tradeoff than other efficient novel view synthesis methods on the NeRF synthetic dataset. The number in the parentheses indicates the model size relative to the baseline NeRF model used in each paper (\textit{best viewed in color}).
    }
    \label{fig:front}
\end{figure}
\begin{abstract}
Recent research explosion on Neural Radiance Field (NeRF) shows the encouraging potential to represent complex scenes with neural networks. One major drawback of NeRF is its prohibitive inference time: Rendering a single pixel requires querying the NeRF network hundreds of times. To resolve it, existing efforts mainly attempt to reduce the number of required sampled points. However, the problem of iterative sampling still exists. On the other hand, Neural \textit{Light} Field (NeLF) presents a more straightforward representation over NeRF in novel view synthesis -- the rendering of a pixel amounts to \textit{one single forward pass} without ray-marching. In this work, we present a \textit{deep residual MLP} network (88 layers) to effectively learn the light field. We show the key to successfully learning such a deep NeLF network is to have sufficient data, for which we transfer the knowledge from a pre-trained NeRF model via data distillation. Extensive experiments on both synthetic and real-world scenes show the merits of our method over other counterpart algorithms. On the synthetic scenes, we achieve $26\sim35\times$ FLOPs reduction (per camera ray) and $28\sim31\times$ runtime speedup, meanwhile delivering \textit{significantly better} ($1.4\sim2.8$ dB average PSNR improvement) rendering quality than NeRF without any customized parallelism requirement.
\end{abstract}

\section{Introduction}
\label{sec:intro}

Inferring the representation of a 3D scene from 2D observations is a fundamental problem in computer graphics and computer vision. Recent research innovations in implicit neural  representations~\cite{chen2019learning,mescheder2019occupancy,park2019deepsdf,takikawa2021neural} and differential neural renders~\cite{mildenhall2020nerf} have remarkably advanced the solutions to this problem. Neural radiance field (NeRF) learned by a simple Multi-Layer Perceptron (MLP) network shows a great potential to store a complex scene into a compact neural network~\cite{mildenhall2020nerf}, thus has inspired plenty of follow-up works~\cite{barron2021mip,dellaert2020neural,li2021neural,yu2021pixelnerf}.

Despite the success of NeRF and its extensions, the drawback is still apparent. The rendering time even for a single pixel is prolonged since the NeRF framework needs to aggregate the radiance of \emph{hundreds of} sampled points via alpha-composition. It requires hundreds of network forwards, thus is prohibitively slow, especially on resource-constrained devices. %
One intuitive solution to the problem is to reduce the model size of NeRF MLP. However, apparent quality degradation of rendered images can be observed (\eg, reducing the network width by only half causes around 0.01 SSIM~\cite{wang2004image} drop in~\cite{rebain2021derf}) while the reduction of inference time is only limited. 
Other research efforts focus on decreasing the number of sampled points~\cite{lindell2021autoint,neff2021donerf}. However, this does not fundamentally resolve the sampling issue. Some work~\cite{neff2021donerf} demands extra depth information for training, which is usually unavailable in most practical cases. 
Thus, a method that only requires \emph{2D images} as input, represents the scene \emph{compactly}, and enjoys a \emph{fast} rendering speed with \emph{high} image quality is highly desired. This paper aims to present such a method that can achieve all the four goals simultaneously by representing the scene as Neural \emph{Light} Field (NeLF) instead of neural \emph{radiance} field. In the neural light field, ray origin and direction are directly mapped into its associated RGB values, avoiding the need of sampling multiple points along the camera ray. Therefore, rendering a pixel requires only one single query, making it much faster than the radiance scene representation.

The idea of NeLF is attractive; however, realizing it for representing \emph{complex real-world} scenes with better quality than NeRF is still challenging.
Our first key technical innovation enabling this is a novel network architecture design for the neural light field network. Specially, we propose a deep (88 layers) residual MLP network with extensive residual MLP blocks. The \emph{deep} network has much greater expressivity than the shallow 
counterparts, thus can represent the light field faithfully. Notably, since the debut of NeRF~\cite{mildenhall2020nerf}, its MLP-based network architecture is inherited with few substantial changes~\cite{barron2021mip,neff2021donerf,rebain2021derf,reiser2021kilonerf}. To our best knowledge, this is the \emph{first} attempt to address the NeRF rendering efficiency issue \emph{from the network design perspective}. Although our network contains more parameters than the original NeRF, we only need \emph{one} single network forward to render the color of a pixel, leading to much faster inference speed than NeRF.

\begin{table}[t]
\centering
\caption{Method comparison between our R2L approach and recent efficient novel view synthesis methods. 
Rendering speedup (measured by FLOPs reduction per ray and wall-time reduction) and representation (Repre.) size are relative to the original NeRF~\cite{mildenhall2020nerf}.
Repre.~size measures the required storage of a neural network or cached files to represent a scene. $\Delta$PSNR refers to the average PSNR improvement (on the NeRF synthetic dataset) over the baseline NeRF used in each paper. Note, ours and~\cite{attal2021learning} are the only two neural \emph{light} field methods here
}
\resizebox{\linewidth}{!}{
\setlength{\tabcolsep}{0.7mm}
\begin{tabular}{lccccc}
\toprule
Method                                  & FLOPs speedup$\uparrow$ & Wall-time speedup$\uparrow$ & Repre.~size$\downarrow$ & Extra design & $\Delta$PSNR (dB)$\uparrow$ \\
\midrule
NeRF~\cite{mildenhall2020nerf}          & $1\times$         &   $1\times$   &   $1\times$           & No    & 0 \\
\hdashline
PlenOctrees~\cite{yu2021plenoctrees}    & -                 & $3000\times$  & $\sim600\times$       & No    & +0.02 \\ %
\hdashline
DONeRF-8~\cite{neff2021donerf}          & $27.60\times$     &  -            & $1.125\times$ & Depth data     & -0.14 \\ %
KiloNeRF~\cite{reiser2021kilonerf}      & $\sim0.6\times$   & $692\times$   & 16.21\x       & Parallelism  & -0.01 \\ %
\hdashline
NSVF~\cite{liu2020neural}               &  -            & $\sim15\times$        & $\sim3.2\times$     & No    & +0.74 \\ %
AutoInt~\cite{lindell2021autoint}       & -             &  $3.22\times$     & $\sim1\times$      & No    & -4.2 \\ %
TermiNeRF~\cite{piala2021terminerf}     & -             & $13.49\times$     & $\sim1\times$     & No    & -0.46 \\ %
RSEN~\cite{attal2021learning}           & -             & $4.86\times^*$    & $1.17\times$      & No    & +0.013 \\ %
\hdashline
Ours                                    & $26\sim35\times$  & $28\sim31\times$ & $4\sim10\times$ & No   & \RE{+1.40} \\
\bottomrule
\end{tabular}}
\label{tab:method_difference}
\end{table}

The major technical problem is how to train the proposed deep residual MLP network. It is well-known that large networks hunger for large sample sizes to curb overfitting~\cite{kearns1994introduction,vapnik2013nature}. We can barely train such a large network using only the original 2D images (which are typically less than $100$ in real-world applications). To tackle this problem, as the second key technical innovation of this paper, we propose to distill the knowledge~\cite{bucilua2006model,hinton2015distilling} from a \emph{pretrained} NeRF model to our network, by rendering pseudo data from random views using the pre-trained NeRF model. We name our method as \textbf{R2L} since we show distilling neural \textbf{R}adiance filed \textbf{to} neural \textbf{L}ight filed is an effective way to obtain a powerful NeLF network for efficient novel view synthesis. Empirically, we evaluate our method on both synthetic and real-world datasets. On the synthetic scenes, we achieve $26\sim35\times$ FLOPs reduction ($28\sim31\times$ wall-time speedup) over the original NeRF with
significantly \emph{higher} rendering quality. Comparison between ours and other efficient novel view synthesis approaches is summarized in Tab.~\ref{tab:method_difference}. 
Overall, our contributions can be summarized into the following aspects:
\begin{itemize}%
    \item Methodologically, we present a brand-new deep residual MLP network aiming for compact neural representation, fast rendering, 
    without extra demand besides 
    2D images, for efficient novel view synthesis. This is the \emph{first} attempt to improve the rendering efficiency via network architecture optimization.
    \item Our network represents complex real-world scenes as neural light fields.
    To resolve the data shortage problem when training the proposed deep MLP network, we propose an effective training strategy by distilling knowledge from a pre-trained NeRF model, which is the key to enabling our method. %
    \item Practically, our approach achieves $26\sim35\times$ FLOPs reduction ($28\sim31\times$ wall-time speedup) over the original NeRF with even better visual quality, which also performs favorably against existing counterpart approaches. %
\end{itemize}

\section{Related Work}
\noindent \bb{Efficient neural scene representation and rendering}.
Since the debut of NeRF~\cite{mildenhall2020nerf}, many follow-up works have been improving its efficiency. One major direction is to skip the empty space and sample more wisely along a camera ray.
NSVF~\cite{liu2020neural} defines a set of voxel-bounded implicit fields organized in a sparse voxel octree structure, which enables skipping empty space in novel view synthesis. %
AutoInt~\cite{lindell2021autoint} improves the rendering efficiency by reducing the number of evaluations along a ray through learned partial integrals.
DeRF~\cite{rebain2021derf} spatially decomposes the scene into Voronoi diagrams, each learned by a small network. They achieve 3 times rendering speedup over NeRF with similar quality. 
Similarly, KiloNeRF~\cite{reiser2021kilonerf} also spatially decomposes the scene, but into thousands of \emph{regular} grids. Each of them is tackled by a tiny MLP network. Their work is similar to ours as a pre-trained NeRF model is also used to generate pseudo samples for training. Differently, KiloNeRF is still a \textit{NeRF}-based method while ours is \textit{NeLF}. Point sampling is still needed in KiloNeRF while our method \textit{roots out} this problem. Besides, KiloNeRF results in \textit{thousands of} small networks, making parallelism more challenging and requiring customized parallelism implementation, while our \textit{single} network can get significant speedup simply using the vanilla PyTorch~\cite{pytorch}.
DONeRF~\cite{neff2021donerf} is proposed recently to reduce sampling through a depth oracle network learned with the ground-truth depth as supervision. It decimates the sampled points from hundreds (\emph{i.e.}, $256$ in NeRF~\cite{mildenhall2020nerf}) to only $4$ to $16$ while maintaining comparable or even better quality. However, the depth oracle network is learned with \emph{ground-truth depth} as target, which is typically unavailable in practice. Our method does not demand it.
Another direction for faster NeRF rendering is to pre-compute and cache the representations per the idea of trading memory
for computational efficiency. FastNeRF~\cite{garbin2021fastnerf} employs a factorized architecture to independently cache the position-dependent and ray direction-dependent outputs and achieves 3000 times faster than the original NeRF at rendering. Baking~\cite{hedman2021baking} precomputes and stores NeRF as sparse neural radiance grid that enables real-time rendering on commodity hardware. We consider this line of works \emph{orthogonal} to ours.

\noindent \bb{Neural light field (NeLF)}. Light fields enjoy a long history as a scene representation in computer vision and graphics~\cite{adelson1991plenoptic,adelson1992single}. Levoy \emph{et al}.~\cite{levoy1996light} and Gortler \emph{et al}.~\cite{gortler1996lumigraph} introduced light fields in computer graphics as 4D scene representation for fast image-based rendering. With them, novel view synthesis can be realized by simply extracting 2D slices in the 4D light field, yet with two major drawbacks. First, they tend to cause considerable storage costs. Second, it is hard to achieve a full 360$^\circ$ representation without concatenating multiple light fields. In the era of deep learning, neural light fields based on convolutional networks have been proposed~\cite{bemana2020x,kalantari2016learning,mildenhall2019local}. One recent neural light field paper is Sitzmann~\emph{et al.}~\cite{sitzmann2021light}. They employ Plücker coordinates to parameterize $360^\circ$ light fields. In order to ensure multi-view consistency, they propose to learn a prior over the 4D light fields in a meta-learning framework. Despite intriguing ideas, their method is only evaluated on toy datasets, not as comparable to NeRF~\cite{mildenhall2020nerf} in representing complex real-world scenes. Another recent NeLF work is RSEN~\cite{attal2021learning}. To tackle the insufficient training data issue, they propose to learn a voxel grid of subdivided \textit{local} light fields instead of the global light field. In their experiments, they also employ a pre-trained NeRF teacher for regularization.
A very recent work~\cite{suhail2022light} proposes a two-stage transformer-based model that can represent view-dependent effects accurately. A concurrent work NeuLF~\cite{liu2022neulf} employs a two-plane parameterization of the light field and uses a vanilla MLP network to learn the NeLF mapping.
Our NeLF network is different from these in that, \textbf{(1)} methodologically, we propose a \emph{deep residual} MLP ($88$ layers) to learn the light field, while these NeLF works still employ the NeRF-like shallow MLP networks (\eg, $6$ layers in \cite{sitzmann2021light}, 8 layers in \cite{attal2021learning}); \textbf{(2)} we propose to leverage a NeRF model to synthesize extra data for training, making our method a bridge from radiance field to light field; \textbf{(3)} thanks to the abundant capacity, our R2L network can achieve better rendering quality (\eg, our method can represent complex real-world scenes against \cite{sitzmann2021light}), or can achieve better efficiency while maintaining the rendering quality (\eg, \cite{attal2021learning} achieves merely around 5\x~speedup~\vs~our 30\x~speedup over the baseline NeRF method).

\noindent \bb{Knowledge distillation (KD)}. The general idea of knowledge distillation is to guide the training of a student model through a larger pre-trained teacher model. Pioneered by Buciluǎ \etal~\cite{bucilua2006model} and later refined by Hinton \etal~\cite{hinton2015distilling} for image classification, knowledge distillation has seen extensive application in vision and language tasks~\cite{chen2017learning,jiao2019tinybert,wang2020collaborative,wang2020knowledge}. Many variants have been proposed regarding the central question in knowledge distillation -- how to define the \emph{knowledge} that is supposed to be transferred from the teacher to the student, examples including output distance~\cite{ba2014deep,hinton2015distilling}, internal feature distance~\cite{romero2014fitnets,wang2020collaborative}, feature map attention~\cite{zagoruyko2016paying}, feature distribution~\cite{passalis2018learning}, activation boundary~\cite{heo2019knowledge}, inter-sample distance relationship~\cite{liu2019knowledge,park2019relational,peng2019correlation,tung2019similarity},  and mutual information~\cite{tian2019contrastive}. The distillation method in this work is to regress the output of the NeRF model with extra data labeled by the teacher (akin to~\cite{ba2014deep,bucilua2006model}), which is the most straightforward way of distillation for the numerical target. Yet we will show this simple scheme can work powerfully to train a deep neural light field network.

\section{Methodology}

\subsection{Background: Neural Radiance Field (NeRF)}
In NeRF~\cite{mildenhall2020nerf},
the 3D scene is implicitly represented by an MLP network, which learns to map the 5D coordinate (spatial location $(x, y, z)$ and viewing direction $(\theta, \phi)$) to the 1D volume density and 3D view-dependent emitted radiance at that spatial location, $F_{\Theta}: \mathbb{R}^{5} \mapsto \mathbb{R}^{4},$
where $F$ refers to an MLP neural network (parameterized by $\Theta$) to represent a scene. For rendering, the classic volume rendering technique~\cite{kajiya1984ray} is adopted in NeRF to obtain the desired color for an oriented ray. Volume rendering is differential thus making NeRF end-to-end trainable by using the captured 2D images as supervision.
For novel view synthesis, given an oriented ray, NeRF first samples several locations along the camera ray, predicts their emitted radiance by querying the MLP network $F_{\Theta}$, and then aggregates the radiance together by alpha composition to output the final color. As sampling at vacuum points contributes nothing to the final color, a sufficient number of sampled points is critical to NeRF's performance so as to cover the worthy locations (such as those near the object surface). However, increased sampling incurs linearly increased query cost of the MLP network. 

\begin{figure*}[t]
\centering
\begin{tabular}{c@{\hspace{0.01\linewidth}}c@{\hspace{0.01\linewidth}}c}
  \includegraphics[width=1\linewidth]{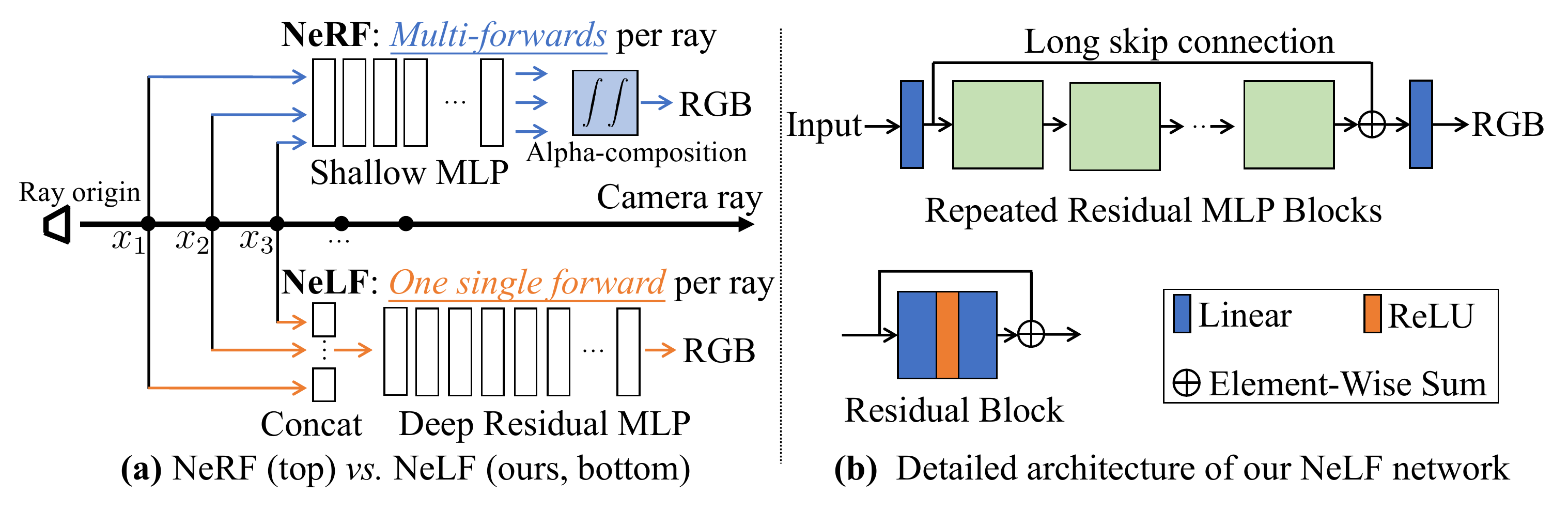}
\end{tabular}
\caption{(a) Comparison between our proposed NeLF network (\emph{Deep Residual MLP}, bottom) and NeRF network (\emph{Shallow MLP}, top).
(b) Detailed architecture of the proposed \emph{deep} light field network, which employs extensive repeated residual MLP blocks. 
}
\label{fig:method_overview}
\end{figure*}

\subsection{R2L: Distilling NeRF to NeLF}
On the other hand, a scene can also be represented as a \emph{light} field instead of \emph{radiance} field, parameterized by a neural network. The network $G_{\phi}$ learns a mapping function directly from a 4D oriented ray to its target 3D RGB, $G_{\phi}: \mathbb{R}^{4} \mapsto \mathbb{R}^{3}.$
NeLF has several attractive advantages over NeRF. \bb{(1)} Methodologically, it is more straightforward for novel view synthesis, in that the output of the NeLF network is already the wanted color, while the output of a NeRF network is the radiance of a sampled point; the desired color has to been obtained through an extra step of ray marching (see Fig.~\ref{fig:method_overview}(a)). 
\bb{(2)} Practically, given the same input ray (origin coordinate and direction), rendering in a light field simply amounts to a \emph{single query} of the light field function. It \emph{fundamentally} obviates the need for point sampling along a ray (which is the speed bottleneck in NeRF~\cite{mildenhall2020nerf}), thus can be orders-of-magnitude faster than NeRF. Despite these intriguing properties, not many successful attempts have crystallized NeLF \emph{with comparable quality to NeRF} up to date. To our best knowledge, only one recent NeLF method~\cite{attal2021learning} achieves comparable quality to NeRF, but its speedup is relatively limited (around 5\x wall-time speedup). In this paper, we propose a novel network architecture to make NeLF as effective as NeRF (meanwhile being much faster). Intuitively, the light field is \emph{harder} to learn than radiance field -- radiance at neighbor space locations does not change dramatically given the radiance field in the physical world is typically continuous; while two neighbor rays can point to starkly different colors because of occlusion. That is, the light field is intrinsically \emph{less smooth} (sharply changing) than the radiance field. To capture the inherently more complex light field, we need a more \emph{powerful} network. Per this idea, the 11-layer MLP network used in NeRF can hardly represent a complex light field by our empirical observation (see Tab.~\ref{tab:ablation_regularization_orignaldata}). We thereby propose to employ a \emph{deep} MLP network to parameterize the above $G$ function. Then, the foremost technical question is how to design the deep network.

\noindent \bb{Network design}. Different from the NeRF network, we propose to employ intensive residual blocks~\cite{resnet} in our network. The resulted network architecture is illustrated in Fig.~\ref{fig:method_overview}(b). Residual connections were shown critical to enable the much greater network depth in \cite{resnet}, which also applies here for learning the light field. The merit of having a \emph{deeper} network will be justified in our experiments (see Fig.~\ref{fig:ablation_studies}(b)). We also study an underperformance case in the supplementary material when the residual connections are \emph{not} used in a deep MLP network. 

Notably, enabling a deep network for neural radiance/light field parameterization is \emph{non-trivial}. Noted by DeRF~\cite{rebain2021derf}, ``\textit{there are diminishing returns in employing larger (deeper and/or wider) networks}''. As a result, notably, most NeRF follow-up works for improving rendering efficiency (\eg,~\cite{rebain2021derf,reiser2021kilonerf,neff2021donerf}) actually inherit the MLP architecture in NeRF with \emph{few} substantial innovations. To our best knowledge, we are the \emph{first} to address the efficiency issue of NeRF \emph{through the network architecture optimization perspective}. Despite the residual structure is not new itself (due to ResNets~\cite{resnet}), its necessity and potential have not been fully recognized and exploited in the NVS task. Our paper is meant to make a step forward in this direction.

\subsection{Synthesize Pseudo Data}
Deep networks hunger for excessive data to be powerful. Unfortunately, this is not the case in novel view synthesis, where a user typically captures fewer than $100$ images. To overcome this problem, we propose to employ a pre-trained NeRF model to synthesize extra data for training. 
This makes our method a bridge from neural \emph{radiance} field to neural \emph{light} field.

We need to decide where to sample to synthesize the pseudo data to avoid unnecessary waste. Specifically, with the original training data (images and their associated camera poses), we know the bounding box of the camera locations and their orientations. Then we \emph{randomly} sample the ray origins $(x_o, y_o, z_o)$ and normalized directions $(x_d, y_d, z_d)$ obeying a uniform distribution $U$ \emph{within the bounding box} to make a 6D input as follows, 
\begin{equation}
    \begin{split}
        x_o \sim U(x_o^{\text{min}}, x_o^{\text{max}}), \; y_o \sim U(y_o^{\text{min}}, y_o^{\text{max}}), \; z_o \sim U(z_o^{\text{min}}, z_o^{\text{max}}), \\
        x_d \sim U(x_d^{\text{min}}, x_d^{\text{max}}), \; y_d \sim U(y_d^{\text{min}}, y_d^{\text{max}}), \; z_d \sim U(z_d^{\text{min}}, z_d^{\text{max}}), \\
    \end{split}
\end{equation}
where the viewing bounding box can be inferred from the training data. An example illustration of the pseudo data origins and directions in our method is shown in our supplementary material. Note, since we can control the generated data, we explicitly demand the pseudo data completely cover the original training data, implying they are in the same domain, which is critical to the performance.

For a trained NeRF model $F_{\Theta^*}$, the target RGB value can be queried as:
\begin{equation}
    (\hat{r}, \hat{g}, \hat{b}) = F_{\Theta^*}(x_o, y_o, z_o, x_d, y_d, z_d),
\end{equation}
where $\Theta^*$ stands for the converged model parameters. Then a slice of training data is simply a vector of these $9$ numbers: $(x_o, y_o, z_o, x_d, y_d, z_d, \hat{r}, \hat{g}, \hat{b})$.
To have  an effective neural light field network $F_{\Theta}$, we feed
abundant pseudo data into the proposed deep R2L network and train it by the MSE loss function,
\begin{equation}
    \mathcal{L} = \text{MSE}(G_{\phi}(x_o, y_o, z_o, x_d, y_d, z_d), (\hat{r}, \hat{g}, \hat{b})).
\label{eq:loss}
\end{equation}

\subsection{Ray Representation and Point Sampling}
It is critical to have a proper representation of a ray in NeLF. %
In this work, we propose a new simple and effective representation -- we concatenate the spatial coordinates of $K$ sampled points along a ray to form an input vector ($3K$-d), fed into the NeLF network.  
\begin{figure}[t]
    \centering
    \includegraphics[width=0.7\linewidth]{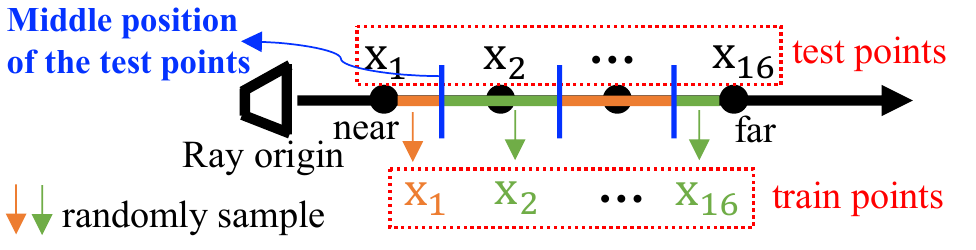}
    \caption{Illustration of the point sampling in training and testing of our method. The \OR{orange} and \GR{green} colors denote the different \textit{segments} of the ray. The \BL{blue} color marks the \textit{start} and \textit{end} points of each segment. Each sampled train point is colored \textit{based on the corresponding segment color}}
    \label{fig:sampling}
\end{figure}
Mathematically, we need at least two points to define a ray. More points will make the representation more precise. In this paper, we choose $K=16$ points (see the ablation of $K$ in Fig.~\ref{fig:ablation_studies}(a)) along a ray. A critical design here is that we expect the network not to overfit the $K$ points but to capture the underlying ray information. Thus, during training the $K$ points are \textit{randomly} sampled along the ray using the stratified sampling (same as NeRF~\cite{mildenhall2020nerf}, see Fig.~\ref{fig:sampling}). This design is critical to generalization. During testing, the $K$ points are evenly spaced. We also tried changing the input to Plücker coordinates for our R2L network (inspired by~\cite{sitzmann2021light}). Our representation achieves \textit{better} test quality than Plücker (PSNR: $29.50$~\vs~$29.08$, scene \texttt{Lego}, W181D88 network, trained with only pseudo data, $200K$ iters).

\subsection{Training with Hard Examples} \label{subsec:hard_examples}
Given that we randomly sample the camera locations and orientations, the rays are likely to point to the trivial parts of a scene (\eg, the white background of a synthetic scene). Also, during training, some easy-to-regress colors will be well-learned early. Feeding these pixels again to the network barely increases its knowledge. We thus propose to tap into the idea of hard examples~\cite{henriques2013beyond,shrivastava2016training}. That is, we want the network to pay more attention to the rays that are harder to regress (typically corresponding to the high-frequency details) during learning. 

Specially, we maintain a \emph{hard example pool}. A \emph{harder} example is defined by a \emph{larger} loss (Eq.~(\ref{eq:loss})). In each iteration, we sort the losses for each sample in a batch in ascending order and add the top $r$ (a pre-defined percentage constant) into the hard example pool. Meanwhile, in each iteration, the same amount $r$ of hard examples are randomly picked out of the pool to augment the training batch. This design can accelerate the network convergence significantly as we will show in the experiments (see Fig.~\ref{fig:ablation_studies}).

\subsection{Implementation Details}
Our R2L can lead to different networks under different FLOPs budgets. In this paper, we mainly have two: 6M and 12M FLOPs (per ray). They result in a bunch of networks: 12M: W256D88, 6M: W181D88, W256D44, W363D22 (W stands for width, D for depth). Obviously, a larger network is expected to perform better,
so W256D88 is used for obtaining better quality; ablation studies will be conducted on the 6M-budget networks since they are faster to train. Following NeRF~\cite{mildenhall2020nerf}, positional encoding~\cite{vaswani2017attention} is used to enrich the input information.

\section{Experiments}
\noindent \bb{Datasets}. We show experiments on the following datasets:
\begin{itemize}[leftmargin=1.5em]
    \item \bb{NeRF datasets}~\cite{mildenhall2020nerf}. 
    We evaluate our method on two datasets: synthetic dataset (Realistic Synthetic $360^{\circ}$) and real-world dataset (Real Forward-Facing). Realistic Synthetic 360$^{\circ}$ contains path-traced images of $8$ objects that exhibit complicated geometry and realistic non-Lambertian materials. $100$ views of each scene are used for training and $200$ for testing, with resolution of $800\times800$. Real Forward-Facing also contains $8$ scenes, captured with a handheld cellphone. There are $20$ to $62$ images for each scene with $1/8$ held out for testing. All images have a resolution of $1008\times756$. 
    \item \bb{DONeRF dataset} includes their synthetic data. Images are rendered using Blender and their Cycles path tracer to render $300$ images for each scene, which are split into train/validation/test sets at a $70\%$, $10\%$, $20\%$ ratio.
\end{itemize}

\noindent \bb{Training settings}. All images in the synthetic dataset are down-sampled by $2\times$ during training and testing. Due to limited space, we defer the full-resolution ($800\times 800$) results to our supplementary material.
The original NeRF model is trained with a batch size of $1,024$ and initial learning rate as $5\times10^{-4}$ (decayed during training) for $200k$ iterations. We synthesize $10k$ images using the pre-trained NeRF model. Our proposed R2L model is trained for $1,000k$ iterations with the same learning rate schedule. The rays in a batch (batch size $98,304$ rays) are randomly sampled from different images so that they do not share the same origin. This is found critical to achieving superior performance. Adam optimizer~\cite{kingma2014adam} is employed for all training. We use PyTorch 1.9~\cite{pytorch}, referring to~\cite{lin2020nerfpytorch}. Experiments are conducted with $8$ NVIDIA V100 GPUs.

\noindent \bb{Comparison methods}. We compare with with the original NeRF~\cite{mildenhall2020nerf} to show that we can achieve significantly better rendering quality while being much faster. Meanwhile, we also compare with DONeRF~\cite{neff2021donerf}, NSVF~\cite{liu2020neural}, and NeX~\cite{wizadwongsa2021nex} since they also target efficient NVS as we do. Other efficient NVS works such as AutoInt~\cite{lindell2021autoint} and X-Fields~\cite{bemana2020x} have been shown less favorable than RSEN~\cite{attal2021learning}. Therefore, we only compare with RSEN~\cite{attal2021learning}. KiloNeRF~\cite{reiser2021kilonerf}, another closely related work apart from RSEN, will also be compared to. Similar to~\cite{attal2021learning}, we do not compare to baking-based methods~\cite{hedman2021baking,yu2021plenoctrees,garbin2021fastnerf}) as they trade memory footprint for speed while our method aims to maintain the compact representation.

\begin{table*}[t]
\centering
\caption{PSNR$\uparrow$, SSIM$\uparrow$, and LPIPS$\downarrow$ (AlexNet~\cite{alexnet} is used for LPIPS) on the NeRF synthetic dataset (Realistic Synthetic 360$^{\circ}$) and real-world dataset (Real Forward-Facing). R2L network: W256D88. $^\dagger$KiloNeRF adopts Empty Space Skipping and Early Ray Termination, so the FLOPs is scene-by-scene; we estimate the average FLOPs based on the description in their paper. The best results are in \RE{red}, second best in \B{blue}}
\resizebox{\linewidth}{!}{
\setlength{\tabcolsep}{1mm}
\begin{tabular}{lcccccccc}
\toprule
\multirow{2}{*}{Method} & \multirow{2}{*}{Storage (MB)} & \multirow{2}{*}{FLOPs (M)} & \multicolumn{3}{c}{Synthetic} & \multicolumn{3}{c}{Real-world} \\
\cline{4-9}
                                            & & & PSNR$\uparrow$ & SSIM$\uparrow$ &LPIPS$\downarrow$ & PSNR$\uparrow$ & SSIM$\uparrow$ &LPIPS$\downarrow$ \\
\midrule
Teacher NeRF~\cite{mildenhall2020nerf}      & 2.4  & 303.82 & 30.47                         & 0.9925       & \BL{0.0391} & \BL{27.68}     & \BL{0.9725}  & \RE{0.0733} \\
Ours-1 (Pseudo)                             & 23.7 & 11.79  & \BL{30.48} (+0.01)            & \BL{0.9939}  & 0.0467 & 27.58 (-0.10) & 0.9722       & 0.0997 \\
Ours-2 (Pseudo+real)                        & 23.7 & 11.79  & \RE{31.87} \textbf{(+1.40)}   & \RE{0.9950}  & \RE{0.0340} & \RE{27.79} \textbf{(+0.11)} & \RE{0.9729} & \BL{0.0968}\\
\hdashline
Teacher NeRF in~\cite{reiser2021kilonerf}   & 2.4  & 303.82               & 31.01         & 0.95 & 0.08 & - & - & - \\ 
KiloNeRF~\cite{reiser2021kilonerf}          & 38.9 & $\sim$500$^\dagger$  & 31.00 (-0.01) & 0.95 & 0.03 & - & - & - \\
\hdashline
Teacher NeRF in~\cite{attal2021learning}    & 4.6 & $\sim$300   & - & - & - & 27.928            & 0.9160 & 0.065 \\
RSEN~\cite{attal2021learning}               & 5.4 & 67.2        & - & - & - & 27.941 (+0.013)   & 0.9161 & 0.060 \\
\bottomrule
\end{tabular}}
\label{tab:psnr_ssim_comparison_blender_llff}
\end{table*}

\begin{figure*}[t]
\centering
\resizebox{0.98\linewidth}{!}{
\begin{tabular}{c@{\hspace{0.005\linewidth}}c@{\hspace{0.005\linewidth}}c@{\hspace{0.005\linewidth}}c@{\hspace{0.005\linewidth}}c@{\hspace{0.005\linewidth}}c@{\hspace{0.005\linewidth}}c@{\hspace{0.005\linewidth}}}
  \includegraphics[width=0.2\linewidth]{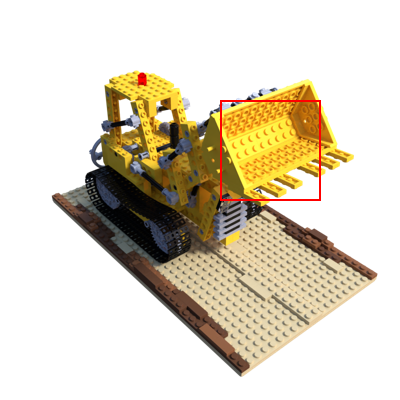} &
  \includegraphics[width=0.2\linewidth]{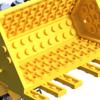} &
  \includegraphics[width=0.2\linewidth]{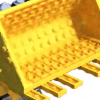} &
  \includegraphics[width=0.2\linewidth]{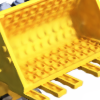} &
  \includegraphics[width=0.2\linewidth]{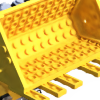} \\
\texttt{Lego} & (a) GT & (b) NeRF~\cite{mildenhall2020nerf} & (c) Ours-1 & (d) Ours-2 \\
  \includegraphics[width=0.2\linewidth]{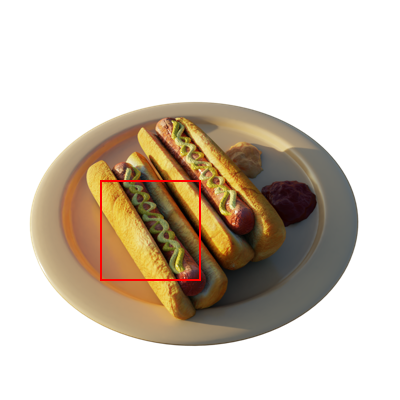} & 
  \includegraphics[width=0.2\linewidth]{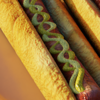} & 
  \includegraphics[width=0.2\linewidth]{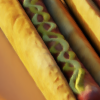} &
  \includegraphics[width=0.2\linewidth]{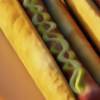} &
  \includegraphics[width=0.2\linewidth]{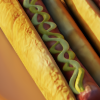} \\
\texttt{Hotdog} & (a) GT & (b) NeRF~\cite{mildenhall2020nerf} & (c) Ours-1 & (d) Ours-2 \\
\end{tabular}}
\caption{Visual comparison between our R2L network (W256D88) and NeRF on the synthetic scene \texttt{Lego} and \texttt{Hotdog}. Ours-1 is trained sorely on pseudo data, ours-2 on pseudo + real data. Please refer to our supplementary material for the visual comparison on the real-world dataset}
\label{fig:comparison_blender}
\end{figure*}

\begin{table*}[t]
\centering
\caption{PSNR$\uparrow$ and FLIP$\downarrow$ comparison on the DONeRF synthetic dataset. All the PSNR and FLIP results except ours and NeRF are directly cited from the DONeRF paper since we are using exactly the same dataset here. Training with pseudo and real data (ours-2) gives us better results.
The best results are in \RE{red}, second best in \B{blue}}
\resizebox{0.9\linewidth}{!}{
\setlength{\tabcolsep}{2mm}
\begin{tabular}{lcccc}
\toprule
Method & Storage (MB) & FLOPs (M) & PSNR$\uparrow$ & FLIP$\downarrow$  \\
\midrule
Teacher NeRF (log+warp)                 & 3.2 & 211.42      & 32.67&0.070 \\
NSVF-large~\cite{liu2020neural}         & 8.3 & 187.52      & 30.01 (-2.66) &0.078 \\
NeX-MLP~\cite{wizadwongsa2021nex}       & 89.0 & 42.71      & 30.55 (-2.12) &0.076 \\
DONeRF-16-noGT~\cite{neff2021donerf}    & 3.6  & 14.29      & 32.25 (-0.42) &0.065 \\
DoNeRF-8~\cite{neff2021donerf}          & 3.6  & \BL{7.66}  & 32.50 (-0.17) &\BL{0.064}   \\               
\hdashline
Ours-1  (Pseduo data)                   & 12.1 & \RE{6.00} & \BL{32.67} (+0.00)   &0.071 \\
Ours-2  (Pseduo + real data)            & 12.1 & \RE{6.00} & \RE{35.45} \textbf{(+2.78)} &\RE{0.047} \\
\bottomrule
\end{tabular}}
\label{tab:psnr_flip_comparison_donerf}
\end{table*}

\begin{figure*}[t]
\centering
\resizebox{0.98\linewidth}{!}{
\begin{tabular}{c@{\hspace{0.005\linewidth}}c@{\hspace{0.005\linewidth}}c@{\hspace{0.005\linewidth}}c@{\hspace{0.005\linewidth}}c@{\hspace{0.005\linewidth}}c@{\hspace{0.005\linewidth}}c@{\hspace{0.005\linewidth}}c@{\hspace{0.005\linewidth}}c@{\hspace{0.005\linewidth}}}
  \includegraphics[width=0.2\linewidth]{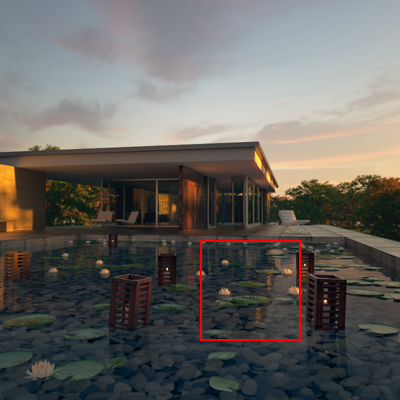} &
  \includegraphics[width=0.2\linewidth]{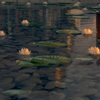} &
  \includegraphics[width=0.2\linewidth]{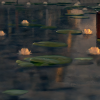} &
  \includegraphics[width=0.2\linewidth]{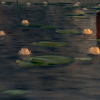} &
  \includegraphics[width=0.2\linewidth]{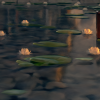} \\
  \texttt{Pavillon} & (a) GT & (b) NeRF & (c) {
  DONeRF-8} & (d) Ours-2 \\
  \includegraphics[width=0.2\linewidth]{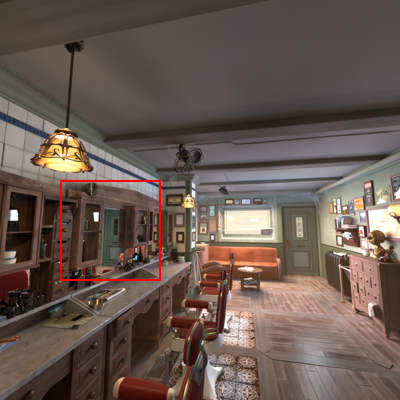} &
  \includegraphics[width=0.2\linewidth]{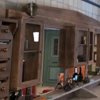} &
  \includegraphics[width=0.2\linewidth]{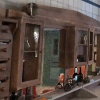} &
  \includegraphics[width=0.2\linewidth]{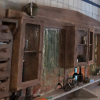} &
  \includegraphics[width=0.2\linewidth]{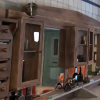} \\
  \texttt{Barbershop} & (a) GT & (b) NeRF & (c) {
  DONeRF-8} & (d) Ours-2 \\
\end{tabular}}
\caption{Visual comparison of ours, NeRF~\cite{mildenhall2020nerf}, DONeRF~\cite{neff2021donerf} on the DONeRF dataset}
\label{fig:comparison_donerf}
\end{figure*}

\subsection{NeRF Synthetic and Real-World Dataset}
The quantitative comparisons (PSNR, SSIM~\cite{wang2004image}, LPIPS~\cite{zhang2018unreasonable}) on the NeRF synthetic and real-world dataset are presented in Tab.~\ref{tab:psnr_ssim_comparison_blender_llff}. Visual comparison is shown in Fig.~\ref{fig:comparison_blender}. \textbf{(1)} Using the pseudo data alone, our R2L network achieves comparable performance to the original ray-marching NeRF model either quantitatively or qualitatively, with only $1/26$ FLOPs. The blurry parts of NeRF results usually also appear on our results, since our model learns from the data generated by the NeRF teacher model. \textbf{(2)} With the original data included for training, our R2L network \textit{significantly} improves the test PSNR \textbf{by $1.40$} over the teacher NeRF model. This means that the performance of our method is \textit{not} upper-bounded by the teacher model. Two primary reasons answer for this remarkable performance. First, our R2L network is \textit{much deeper} than the NeRF network, which bestows a much greater capacity to represent scenes with fine-grained details. Second, we propose \textit{hard-example training} (Sec.~\ref{subsec:hard_examples}), which makes the network focus more on regressing the fine-grained details.
\textbf{(3)} For the related works KiloNeRF and RSEN, their baseline NeRF models have different PSNRs due to different settings, so the PSNR results cannot be directly compared. Instead, we compare the \textit{PSNR change} over the baseline NeRFs. KiloNeRF gets $0.01$ dB PSNR drop~\vs~ours $1.40$ dB PSNR boost. RSEN improves the PSNR on the much more challenging real-world dataset marginally (by $0.013$ dB). In comparison, our improvement is more significant ($0.11$ dB) with much fewer FLOPs.

\begin{table*}[t]
\centering
\caption{Average time (s) comparison among our R2L network (W181D88), DONeRF, and NeRF. The benchmark is conducted under the \emph{same} hardware and software. The speedup of ours and DONeRF is relative to the running time of NeRF. Results are averaged by 60 frames}
\resizebox{\linewidth}{!}{
\setlength{\tabcolsep}{1.2mm}
\begin{tabular}{l|cccc}
\toprule
Method & FLOPs (M) & GeForce 2080Ti & Tesla V100 & CPU \\
\midrule
NeRF                & 211.42                    & 5.9343                        & 4.9902    & 142.2612 \\
DONeRF-16           & 14.29 (14.79$\times$)     & 0.4162 (14.26$\times$)        & 0.3524 (14.16$\times$)         & 9.9344 (14.32$\times$) \\
Ours                & \bb{6.00 (35.24$\times$)} & \bb{0.2103 (28.22$\times$)}   & \bb{0.1629 (30.63$\times$)} & \bb{5.0198 (28.34$\times$)} \\
\bottomrule
\end{tabular}}
\label{tab:actual_speed_comparison}
\end{table*}

\begin{table*}[t]
\centering
\caption{Ablation study of different network and data schemes when learning a light field. Scene: \texttt{Lego}. All models are trained for $200k$ iterations. Note, the train PSNR of our method is lower than test PSNR because we use the hard examples (Sec.~\ref{subsec:hard_examples})~\ie, examples with small PSNR, for training.}
\resizebox{\linewidth}{!}{
\setlength{\tabcolsep}{1mm}
\begin{tabular}{lcccc}
\toprule
Network & Data & Train PSNR (dB) & Test PSNR (dB) \\
\midrule
NeRF~\cite{mildenhall2020nerf} & Original ($0.1k$ imgs) & 25.61 & 19.81 \\
NeRF+dropout~\cite{srivastava2014dropout} &  Original ($0.1k$ imgs) & 25.56 & 19.83 \\
NeRF+BN~\cite{ioffe2015batch} & Original ($0.1k$ imgs) & 25.43 & 19.76 \\
NeRF~\cite{mildenhall2020nerf} & Pseudo ($10k$ imgs) & 23.82 & 26.67 \\
\hdashline
R2L (W181D88) & Pseudo ($10k$ imgs) & 28.38 & \BL{29.50} \\
R2L (W181D88) & Pseudo + Original ($10.1k$ imgs) & 29.85 & \RE{30.09} \\
\bottomrule
\end{tabular}}
\label{tab:ablation_regularization_orignaldata}
\end{table*}

\subsection{DONeRF Synthetic Dataset}
DONeRF~\cite{neff2021donerf} achieves fast rendering using \emph{ground-truth depth} for training. However, the ground-truth depth is \emph{not} available in most practical cases. As a remedy, they propose to use a pre-trained NeRF model to estimate depth as a proxy for the ground-truth depth. The approach of DONeRF without ground-truth depth (\emph{e.g.}, DONeRF-16-noGT) is very relevant to ours. Thus, we compare with it using the synthetic dataset collected by the DONeRF paper.
The quantitative results (PSNR and FLIP~\cite{andersson2020flip}) are presented in Tab.~\ref{tab:psnr_flip_comparison_donerf}. \textbf{(1)}~Trained purely with pseudo data, our method already outperforms DONeRF-16-noGT and DONeRF-8 (which even demands the ground-truth depth as input). \textbf{(2)} Similar to the case (Tab.~\ref{tab:psnr_ssim_comparison_blender_llff}) on the NeRF synthetic dataset, including the original real images for training significantly boosts the performance by $2.78$ dB. 

Visual results in Fig.~\ref{fig:comparison_donerf} show that our method delivers better visual quality than the baseline NeRF. On the scene \texttt{Pavillon} and \texttt{Barbershop}, our R2L network achieves \emph{better} rendering quality than DONeRF-8 despite not using the ground-truth depth. Particularly note the reflection surfaces (\eg, water in \texttt{Pavillon} and mirror in \texttt{Barbershop}), DONeRF cannot learn the reflection surfaces well because the ground-truth depth does not apply to the depth in the reflections, while our method (and the original NeRF) still performs well.

\noindent \bb{Actual speed comparison}. We further report the benchmark results of wall-time speed in  Tab.~\ref{tab:actual_speed_comparison} to demonstrate the FLOPs reduction is well-aligned with actual speedup. Our R2L network (W181D88) is $28\sim31\times$ faster than NeRF and $2\times$ faster than DONeRF-16-noGT.

\begin{figure*}[t]
\centering
\resizebox{0.999\linewidth}{!}{
\begin{tabular}{c@{\hspace{0.01\linewidth}}c@{\hspace{0.01\linewidth}}c}
    \includegraphics[width=0.49\linewidth]{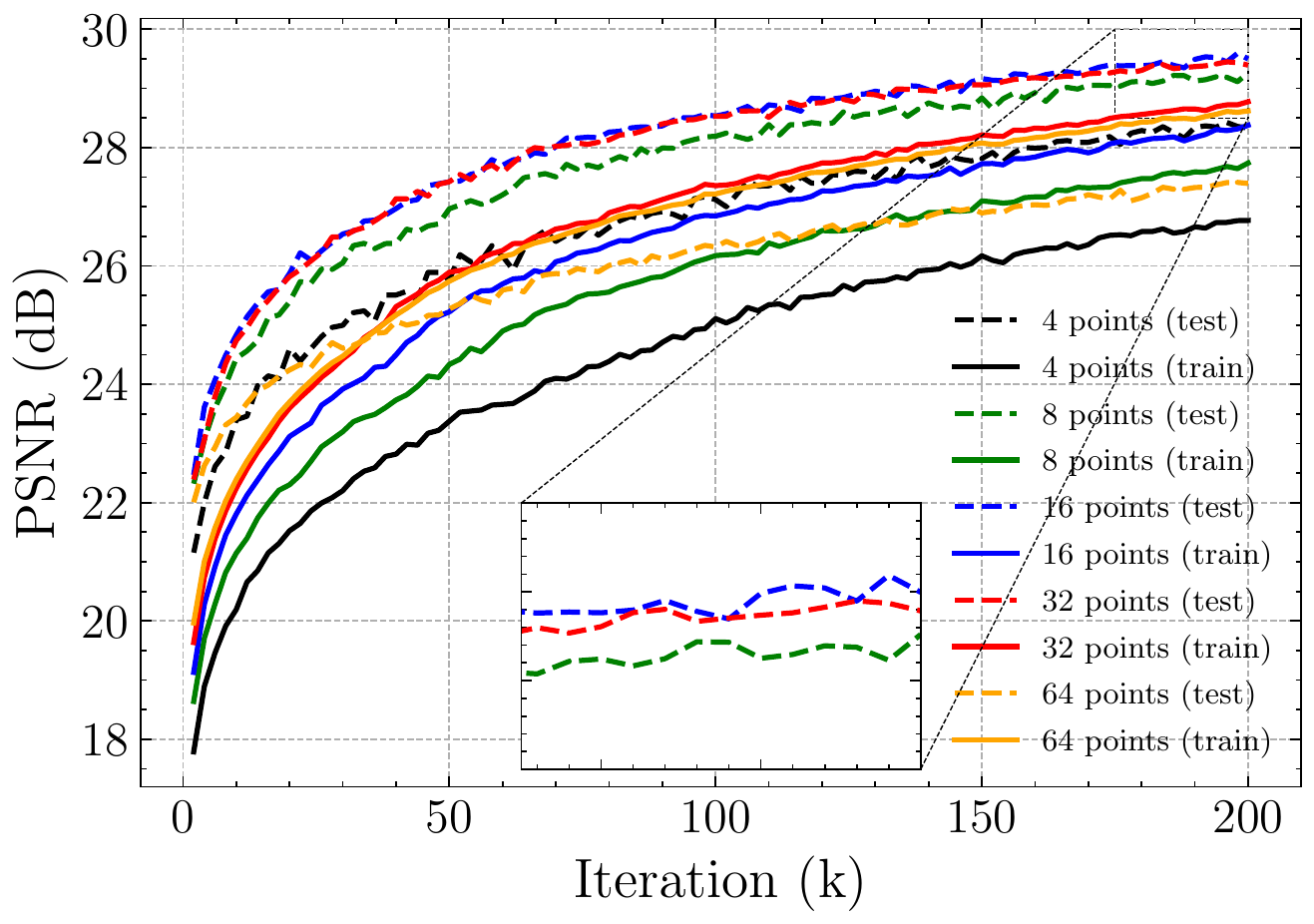} &
    \includegraphics[width=0.49\linewidth]{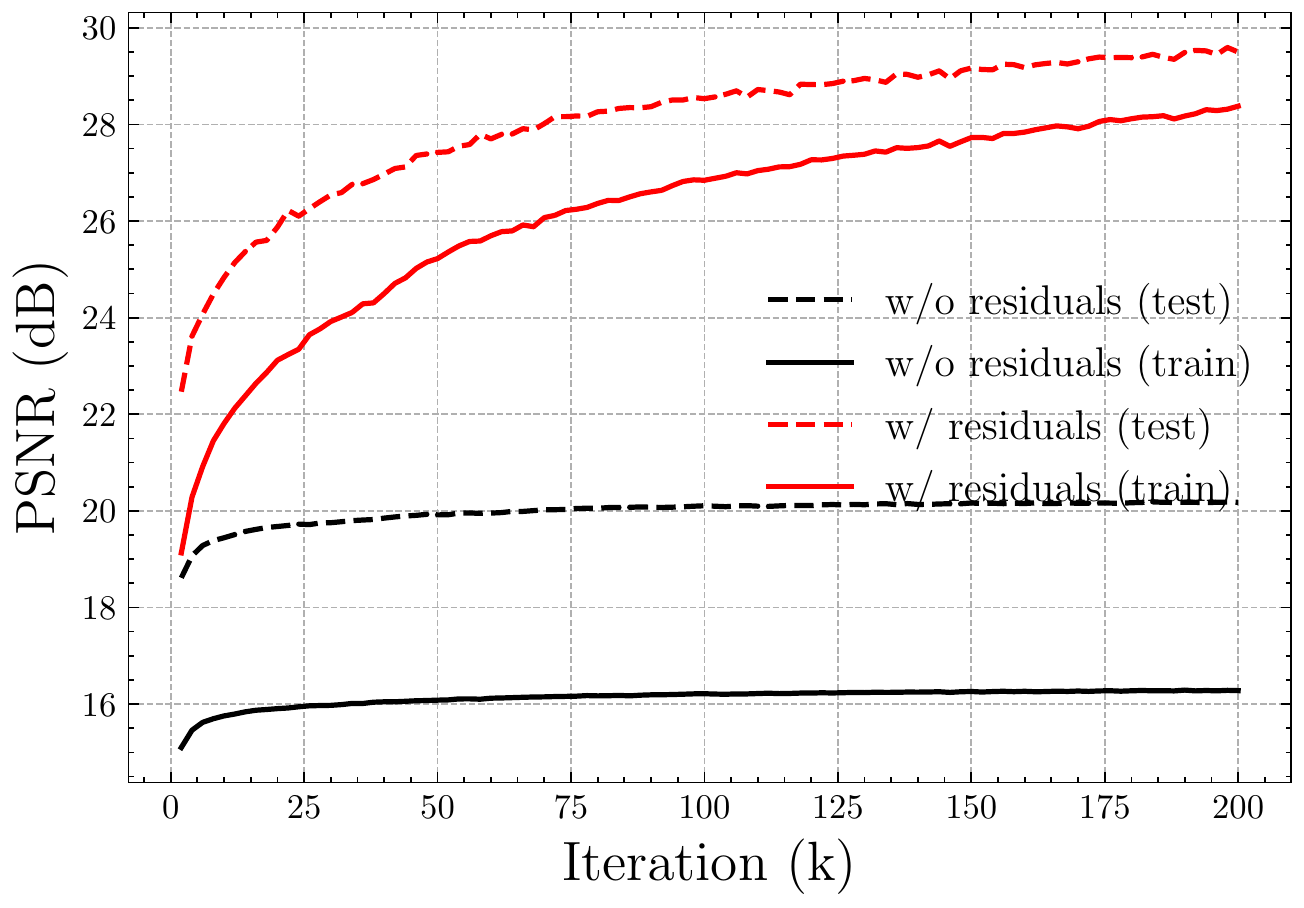} \\
    (a) Number of sampled points & (b) With~\vs~without residuals \\
    \includegraphics[width=0.49\linewidth]{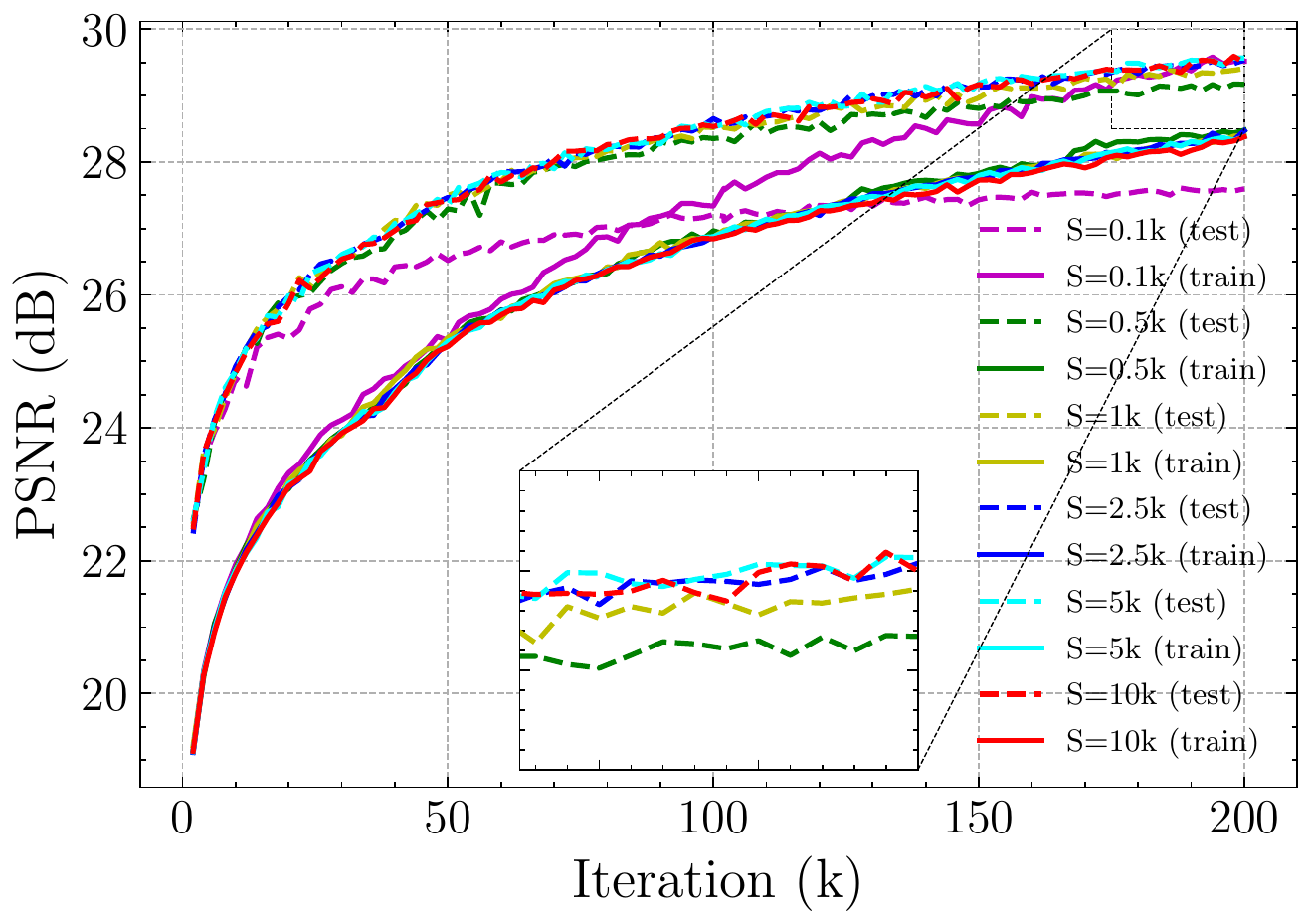} &
    \includegraphics[width=0.49\linewidth]{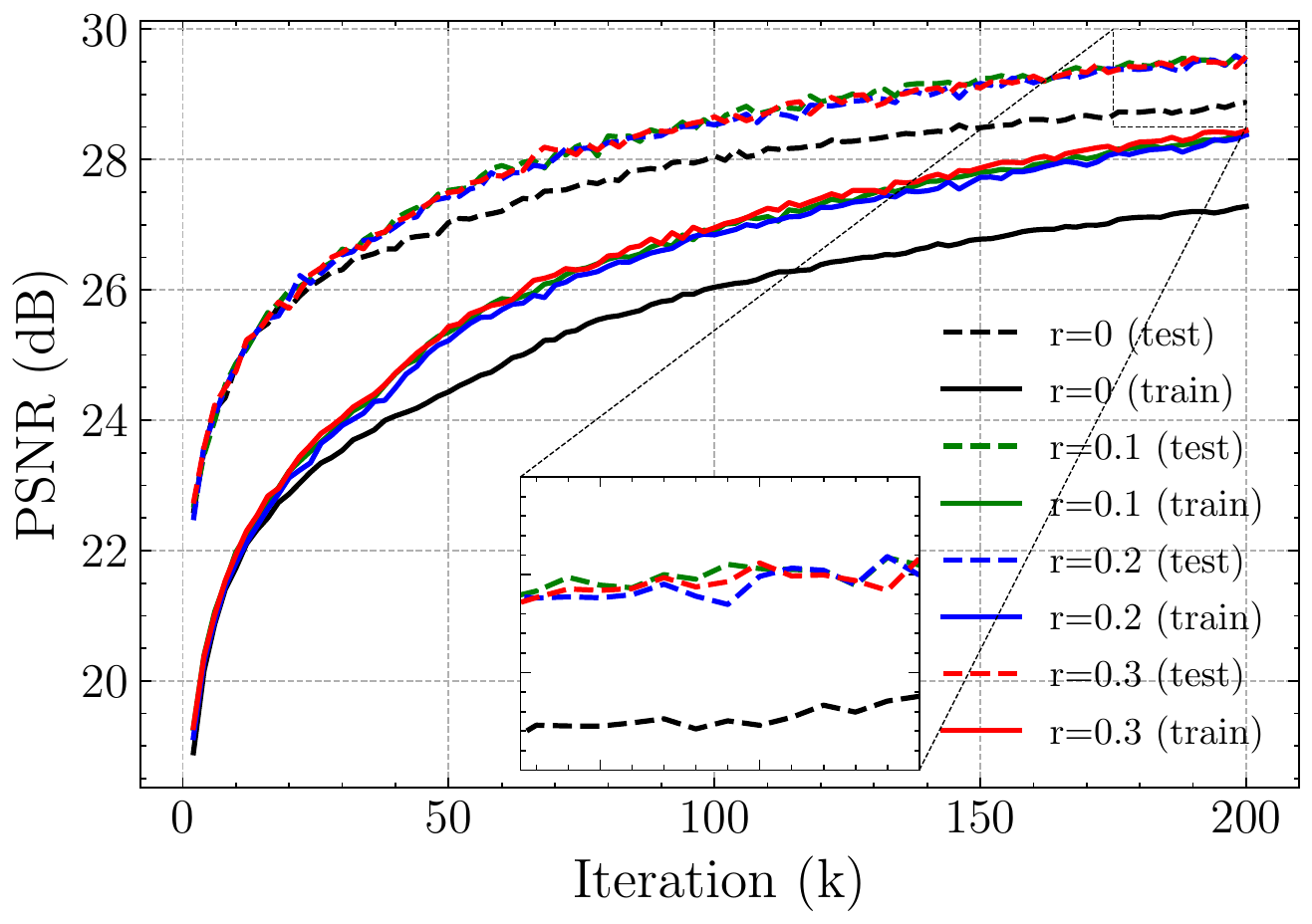} \\
    (c) Pseudo sample size & (d) Hard example ratio \\
\end{tabular}}
\caption{Ablation studies. All networks are trained for $200k$ iterations, scene: \texttt{Lego}. Test PSNRs are plotted with dashed lines; train PSNRs are plotted with solid lines. \bb{(a)} PSNR comparison of different sampled points in our R2L network (W181D88). Default: 16 points (\textcolor{blue}{blue} lines) \bb{(b)} PSNR comparison between two network designs: using residuals or not for our R2L network.  \bb{(c)} PSNR comparison under different pseudo sample sizes. Default: $S=10k$. \bb{(d)} PSNR comparison under different hard example ratios $r \in \{0, 0.1, 0.2, 0.3\}$. Default: $r=0.2$}
\label{fig:ablation_studies}
\end{figure*}

\subsection{Ablation Study}
\noindent \bb{\textit{More} data and \textit{deep} network are critical}. Tab.~\ref{tab:ablation_regularization_orignaldata} shows the results of using the original 11-layer NeRF network to learn a light field on scene \texttt{Lego}. \textbf{(1)} Because of the severely insufficient data (only $0.1k$ training images), the network overfits to the training data with only $19.81$ test PSNR. Note, this overfitting cannot be resolved by common regularization techniques like dropout~\cite{srivastava2014dropout} and BN~\cite{ioffe2015batch}. Only when the data size is greatly inflated (with pseudo data) from $0.1k$ to $10k$, can we see a significant test PSNR improvement (from $19.81$ to $26.67$). This shows the (abundant) pseudo data is indispensable. \textbf{(2)} Compare our R2L to NeRF at the same setting of $10k$ pseudo images, our network design improves test PSNR by around $3$ (from $26.67$ to $29.50$), which is a significant boost in terms of rendering quality. This justifies the necessity of our \textit{deep} network design. Another reason encouraging us to use deep networks is that we empirically find trading width for depth under the same FLOPs budget can consistently lead to performance gains (see our supplementary material).
 
\noindent \bb{Ablation of residuals in our R2L network}. Although the original NeRF network also employs skip connections (to add ray directions as input), it can hardly be considered as a typical residual network~\cite{resnet} in fact, as they do not use residuals in the internal layers. In comparison, we promote employing extensive residual blocks in the internal layers. Its necessity is justified by Fig.~\ref{fig:ablation_studies}(b). As seen, without residuals, the network is barely trainable. 

\noindent \bb{Ablation of pseudo sample size}.
The effect of pseudo sample size is of particular interest. As shown in Fig.~\ref{fig:ablation_studies}(c), $100$ images (see $S=0.1k$) are not enough to train our deep R2L network -- note the test PSNR saturates early at around 50$k$ iterations while its train PSNR keeps arising sharply. This is a typical case of overfitting, caused by the over-parameterized model not being fed with enough data. In contrast, with more data (see the cases of $S \ge 0.5k$), the train PSNR is held down and the test PSNR keeps arising. We observe no significant improvement starting from around $5k$ images.

\noindent \bb{Ablation of hard example ratio}. Here we vary the hard example ratio $r$ and see how it affects the performance. To make a fair comparison, we keep the training batch size always the same ($98,304$ rays per batch) when varying $r$. 
As shown in Fig.~\ref{fig:ablation_studies}(d), using hard examples in each batch significantly improves the network learning in either train PSNR (\ie, better optimization) or test PSNR (\ie, better generalization) against the case of $r=0$. There is no significant difference between hard example ratio $r=0.1$, $0.2$, and $0.3$. In our experiments, we simply use a setting as $r=0.2$.

\section{Conclusion}
We present the first \textit{deep} neural light field network that can represent complex synthetic and real-world scenes. Starkly different from existing NeRF-like MLP networks, our R2L network is featured by an unprecedented depth and extensive residual blocks. We show the key to training such a deep network is abundant data, while the original captured images are barely sufficient. To resolve this, we propose to adopt a pre-trained NeRF model to synthesize excessive pseudo samples. With them, our proposed neural light field network achieves more than $26\sim35\times$ FLOPs reduction and $28\sim31\times$ wall-time acceleration on the NeRF synthetic dataset, with rendering quality improved significantly.

\clearpage
\bibliographystyle{splncs04}
\bibliography{references}
\end{document}